# A Comparative Study of Meta-heuristic Algorithms for Solving Quadratic Assignment Problem

Gamal Abd El-Nasser A. Said

Computer Science Department
Faculty of Computer & Information
Sciences, Ain Shams University
Cairo, Egypt

Abeer M. Mahmoud

Computer Science Department
Faculty of Computer & Information
Sciences, Ain Shams University
Cairo, Egypt

El-Sayed M. El-Horbaty

Computer Science Department
Faculty of Computer & Information
Sciences, Ain Shams University
Cairo, Egypt

*Abstract*—**Quadratic Assignment Problem (QAP) is an NP-hard combinatorial optimization problem, therefore, solving the QAP requires applying one or more of the meta-heuristic algorithms. This paper presents a comparative study between Meta-heuristic algorithms: Genetic Algorithm, Tabu Search, and Simulated annealing for solving a real-life (QAP) and analyze their performance in terms of both runtime efficiency and solution quality. The results show that Genetic Algorithm has a better solution quality while Tabu Search has a faster execution time in comparison with other Meta-heuristic algorithms for solving QAP.**

## I. INTRODUCTION

Optimization problems arise in various disciplines such as engineering design, manufacturing system, economics etc. thus in view of the practical utility of optimization problems there is a need for efficient and robust computational algorithms which can solve optimization problems arising in different fields. Several NP-hard combinatorial optimization problems, such as the traveling salesman problem, and yard management of container terminals can be modeled as QAPs..

Optimization is a process that finds a best, or optimal, solution for a problem. An optimization problem is defined as: Finding values of the variables that minimize or maximize the objective function while satisfying the constraints. The Optimization problems are centered on three factors: (1) an objective function which is to be minimized or maximized. (2) A set of unknowns or variables that affect the objective function. (3) A set of constraints that allow the unknowns to take on certain values but exclude others.

In most optimization problems there is more than one local solution. Therefore, it becomes very important to choose a good optimization method that will not be greedy and look only in the neighborhood of the best solution; because this will mislead the search process and leave it stuck at a local solution. However, the optimization algorithm should have a mechanism to balance between local and global search. There are multiple methods used to solve optimization problems of both the mathematical and combinatorial types. In fact, if the optimization problem is difficult or if the search space is large, it will become difficult to solve the optimization problem by using conventional mathematics.

For this reason, many meta-heuristic optimization methods have been developed to solve such difficult optimization problems [13].

Combinatorial generally means that the state space is discrete. Combinatorial optimization is widely applied in a number of areas nowadays. Combinatorial optimization problems (COP) are those problems that have a finite set of possible solutions. The best way to solve a combinatorial optimization problem is to check all the feasible solutions in the search space. However, checking all the feasible solutions is not always possible, especially when the search space is large. Thus, many Meta-heuristic algorithms have been devised and modified to solve these problems. The Meta-heuristic approaches are not guaranteed to find the optimal solution since they evaluate only a subset of the feasible solutions, but they try to explore different areas in the search space in a smart way to get a near-optimal solution in less cost and time [11].

In this paper we focus on combinatorial optimization problem, namely the Quadratic Assignment Problem. (QAP) is one of the most difficult NP-hard combinatorial optimization problems, so, to practically solve the QAP one has to apply Meta-heuristic algorithms which find very high quality solutions in short computation time[15].

The rest of this paper is organized as follows: A brief description of QAP is given in section II. Section III provides a brief overview of related work for comparison between different Meta-heuristic algorithms for solving combinatorial problems. The Meta-heuristic algorithms are described in section IV. This section is further subdivided into three subsections namely GA, TS, and SA. Section V, include the experimental results. Our conclusions and future work are given in section VI.

## II. QUADRATIC ASSIGNMENT PROBLEM

QAP is one of the most difficult NP-hard combinatorial optimization problems; there are a set of n facilities and a set of n locations. For each pair of locations, a distance is specified and for each pair of facilities a weight or flow is specified. The problem is to assign all facilities to different locations with the aim of minimizing the sum of the distances multiplied by the corresponding flows. (QAP) is formulated as follows:-

The following notation is used in formulation of QAP

n     total number of facilities and locations

$f_{ik}$   flow of material from facility I to facility k

$d_{jl}$   distance from location j to location l

The objective function minimizes the total distances and flows between facilities

$$\min \quad f(x) = \sum_{i=1}^{n}\sum_{j=1}^{n}\sum_{k=1}^{n}\sum_{l=1}^{n} f_{ik} d_{jl} x_{ij} x_{kl}$$

$$\text{s.t} \quad \sum_{j=1}^{n} x_{ij} = 1,$$

$$\sum_{i=1}^{n} x_{ij} = 1,$$

where

$$x_{ij} = \begin{cases} 1, & \text{if facility i is assigned to location j} \\ 0, & \text{otherwise} \end{cases}$$

The constraints ensure that each facility i is assigned to exactly one location j and each location j has exactly one facility which assigned to it [16].

### III. RELATED WORK

Many Meta-heuristic algorithms have been proposed by researchers to find optimal or near optimal solutions for the QAP such as Genetic Algorithm [1], Tabu Search [3] and Simulated Annealing [15]. Also Many researchers presented comparison study between different Meta-heuristic algorithms for solving combinatorial problems [2,5,11].

John Silberholz and Bruce Golden [2], compared Meta-heuristic algorithms in terms of both solution quality and runtime, Their conclusions show that good techniques in solution quality and runtime comparisons will ensure fair and meaningful comparisons are carried out between Meta-heuristic algorithms, producing the most meaningful and unbiased results possible.

Comparison between simulated annealing and genetic algorithm for solving the Travelling Salesman Problem was done by Adewole et al. [4], where they have compared the performance of SA and GA. Their results show that Simulated Annealing runs faster than Genetic Algorithm and runtime of Genetic Algorithm increases exponentially with number of cities. However, in terms of solution quality Genetic Algorithm is better than Simulated Annealing.

Bajeh et al. [5] compared Genetic Algorithm and Tabu Search approaches to solve scheduling problems. The results show that TS can produce better solution, with less computing time, than those produced by GA. However, GA can produce several different near optimal solutions at the same time because of its holds the whole generation of chromosomes which may not originate from the same parents.

Marvin et al. [6] compared the relative performance of Tabu Search (TS), Simulated Annealing (SA) and Genetic Algorithms (GA) on various types of FLP under time-limited, solution-limited, and unrestricted conditions. The results indicate that TS shows very good performance in most cases. The performance of SA and GA are more partial to problem type and the criterion used.

In Karimi et al. [7] Meta-heuristic methods such as SA, TS, and PSO are presented; the research is dedicated to compare the relative percentage deviation of these solution qualities from the best known solution which is introduced in QAPLIB. The results show that TS is the most excellent method in computational time.

Paul [8] compared the performance of tabu search and simulated annealing heuristics for the quadratic assignment problem. The results shows that for a number of varied problem instances, SA performs better for higher quality targets while TS performs better for lower quality targets.

This paper presents Genetic algorithm (GA), Tabu search (TS) and simulated annealing (SA) for solving real life Quadratic Assignment Problem. The analysis of the obtained results in terms of both runtime efficiency and solution quality show the performance of each algorithm and show comparison on their effectiveness in finding the optimal solution for real life QAP.

### IV. META-HEURISTIC ALGORITHMS

A Meta-heuristic is formally defined as an iterative generation process which guides a subordinate heuristic by combining intelligently different concepts for exploring and exploiting the search space, learning strategies are used to structure information in order to find efficiently near-optimal solutions. Meta-heuristic algorithms are among these approximate techniques which can be used to solve complex problems.

Most widely known Meta-heuristic algorithms are Genetic algorithm (GA), simulated annealing (SA) and Tabu search (TS). Genetic algorithm (GA) emulate the evolutionary process in nature, whereas tabu search (TS) exploits the memory structure in living beings, simulated annealing (SA) imitates the annealing process in crystalline solids [2].

#### A. Genetic algorithm

Genetic Algorithm is a Meta-heuristic algorithm that aims to find solutions to NP-hard problems. The basic idea of Genetic Algorithms is to first generate an initial population randomly which consist of individual solution to the problem called Chromosomes, and then evolve this population after a number of iterations called Generations. During each generation, each chromosome is evaluated, using some measure of fitness. To create the next generation, new chromosomes, called offspring, are formed by either merging two chromosomes from current generation using a crossover operator or modifying a chromosome using a mutation operator. A new generation is formed by selection, according to the fitness values, some of the parents and offspring, and rejecting others so as to keep the population size constant. Fitter chromosomes have higher probabilities of being selected. After several generations, the algorithms converge to the best chromosome, which hopefully represents the optimum or suboptimal solution to the problem [12]. The process of GA can be represented as follows:

*Step 1 Generate initial population.*

*Step 2 Evaluate populations.*

*Step 3 Apply Crossover to create offspring.*

*Step 4 Apply Mutation to offspring.*

*Step5 Select parents and offspring to form the new population for the next generation.*

*Step 6 If termination condition is met finish, otherwise go to Step2*

### B. Tabu Search

Tabu search is the technique that keeps track of the regions of the solution space that have already been searched in order to avoid repeating the search near these areas [8]. It uses from a random initial solution and successively moves to one of the neighbors of the current solution. The difference of tabu search from other Meta-heuristic approaches is based on the notion of tabu list, which is a special short term memory. That is composed of previously visited solutions that include prohibited moves. In fact, short term memory stores only some of the attributes of solutions instead of whole solution. So it gives no permission to revisited solutions and then avoids cycling and being stuck in local optima.

During the local search only those moves that are not tabu will be examined if the tabu move does not satisfy the predefined aspiration criteria. These aspiration criteria are used because the attributes in the tabu list may also be shared by unvisited good quality solutions. A common aspiration criterion is better fitness, i.e. the tabu status of a move in the tabu list is overridden if the move produces a better solution [2, 3]. The process of TS can be represented as follows:

*Step 1 Generate initial solution x.*

*Step 2 Initialize the Tabu List.*

*Step 3 While set of candidate solutions X″ is not complete.*

*Step 3.1 Generate candidate solution x″ from current solution x*

*Step 3.2 Add x″ to X″ only if x″ is not tabu or if at least one Aspiration Criterion is satisfied.*

*Step 4 Select the best candidate solution x* in X″.*

*Step 5 If fitness(x*) > fitness(x) then x = x*.*

*Step 6 Update Tabu List and Aspiration Criteria*

*Step 7 If termination condition met finish, otherwise go to Step 3.*

### C. Simulated annealing

Simulated Annealing is an early Meta-heuristic algorithm originating from an analogy of how an optimal atom configuration is found in statistical mechanics. It uses temperature as an explicit strategy to guide the search. In Simulated Annealing, the solution space is usually explored by taking random tries. The Simulated Annealing procedure randomly generates a large number of possible solutions, keeping both good and bad solutions.

As the simulation progresses, the requirements for replacing an existing solution or staying in the pool becomes stricter and stricter, mimicking the slow cooling of metallic annealing. Eventually, the process yields a small set of optimal solutions. Simulated Annealing advantage over other methods is its ability to obviate being trapped in local minima.

This means that the algorithm does not always reject changes that decrease the objective function or changes that increase the objective function according to its probability function: $p = e^{\Delta f/T}$ Where T is the control parameter (analogy to temperature) and $\Delta f$ is the variation in the objective function [4,15]. The process of SA can be represented as follows:

*Step 1 Compute randomly next position .*

*Step2 Determine the difference between the next position and current position, call this different delta .*

*Step3 If delta < 0, the assign the next position to the current position .*

*Step4 If delta > 0, then compute the probability of accepting the random next position .*

*Step5 If the probability is < the e^(-delta / temperature), then assign the next position to the current position .*

*Step 6 Decrease temperature by a factor of alpha .*

*Step7 Loop to step 1 until temperature is not greater than epsilon*

## V. EXPERIMENTS AND RESULTS

Experimental results were run on a Laptop with the following configurations: i3 CPU 2.4 GHZ, 4.0 GB RAM, Windows 7. This test was conducted with GA, TS and SA algorithms. Comparison of the algorithms is based on solution quality and execution time for real life QAP, in the experiment, we analyze the solution quality and run time for solving QAP using instances presented in QAPLIB site [18].

QAPLIB problems are classified to four classes. (i) Unstructured, randomly generated instances. (ii) Grid- based distance matrix instances (iii) Real-life instances (iv) Real-life like instances.

These groups are taken from the study of Ramkumar et al. (2009) [17]. Our experiment for Real-life instances class which have Best Known Quality Solution as shown in the Table1.

TABLE I.     QAP AT DIFFERENT PROBLEMS SIZE

| Problem Name | Problem Size | Best Known Quality |
|---|---|---|
| bur26h | 26 | 7098658 |
| chr12c | 12 | 11156 |
| Chr15a | 15 | 9896 |
| Esc128 | 128 | 64 |
| esc16i | 16 | 14 |
| esc32h | 32 | 438 |
| esc64a | 64 | 128 |
| had12 | 12 | 1652 |
| had14 | 14 | 2724 |
| had20 | 20 | 6922 |
| kra30b | 30 | 91420 |
| ste36a | 36 | 9526 |

The obtained Best quality solutions for each algorithm of Meta-heuristic algorithms are compared with QAPLIB Best Known Quality solutions. For each problem instance we execute a series of runs for various parameters.

Figure.1 shows an example of solving QAP namely ste36a. The figure shows the best Quality solution and the best known Quality solution of the problem ste36a by tabu search algorithm.

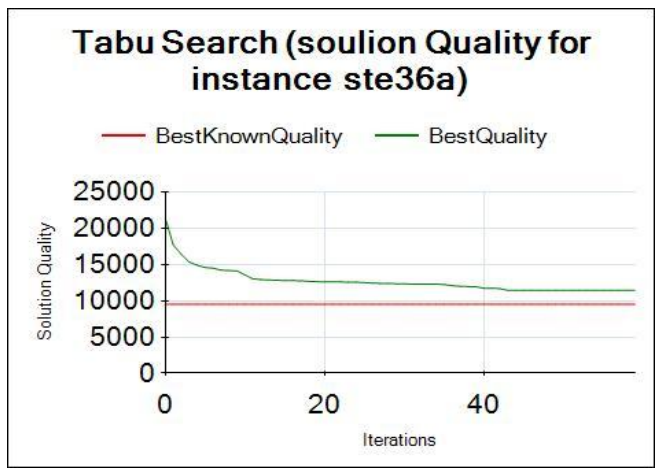

Fig. 1.   solution quality of instance ste36a using tabu search algorithm

The results shown in table 2 show the relative differences of the solution quality for real life quadratic assignment problems by each algorithm for different problems size.

Relative difference represents the difference between algorithm best quality solution and the best known quality solution of the problem in percent. The difference value is calculated in the following way

Relative difference= ((Best Quality − Best Known Quality) / Best Known Quality) * 100%

TABLE II.     RELATIVE DIFFERENCE OF THE SOLUTION QUALITY FOR QAP AT DIFFERENT PROBLEMS SIZE

| Problem Nam | GA Diff% | TS Diff% | SA Diff% |
|---|---|---|---|
| bur26h | 0.84% | 1.51% | 0.96% |
| chr12c | 9% | 25.91% | 16.22% |
| Chr15a | 27.21% | 77.66% | 32.95% |
| Esc128 | 91.67% | 73.26% | 60.94% |
| esc16i | 0% | 3% | 0% |
| esc32h | 7.23% | 10.13% | 3.88% |
| esc64a | 1% | 1.07% | 0% |
| had12 | 0.61% | 1.70% | 0.87% |
| had14 | 0% | 1.31% | 0% |
| had20 | 0.93% | 2.23% | 0.82% |
| kra30b | 6.09% | 14.38% | 14.44% |
| ste36a | 36.70% | 31.81% | 38.52% |

As for the execution time, in table 3 we compare the execution time by each algorithm for QAP at different problems size. Execution time in the format (minutes: seconds. tenths of seconds).

TABLE III.     SOLUTION EXECUTION TIME FOR QAP AT DIFFERENT PROBLEMS SIZE

| Problem Name | Execution Time | Execution Time | Execution Time |
|---|---|---|---|
|  | GA | TS | SA |
| bur26h | 00:10.4 | 00:00.4 | 00:02.7 |
| chr12c | 00:08.9 | 00:00.1 | 00:02.3 |
| Chr15a | 00:08.6 | 00:00.2 | 00:02.5 |
| Esc128 | 00:33.6 | 00:17.8 | 00:13.7 |
| esc16i | 00:10.0 | 00:00.1 | 00:03.6 |
| esc32h | 00:10.2 | 00:00.4 | 00:03.3 |
| esc64a | 00:15.4 | 00:01.6 | 00:04.9 |
| had12 | 00:08.6 | 00:00.2 | 00:02.4 |
| had14 | 00:08.7 | 00:00.2 | 00:02.4 |
| had20 | 00:09.0 | 00:00.3 | 00:02.5 |
| kra30b | 00:10.2 | 00:00.5 | 00:02.8 |
| ste36a | 00:10.8 | 00:00.6 | 00:03.0 |

Figure.2  shows the relative percentage deviation (Relative difference) of the solution quality for different problems size for GA,TS and GA algorithms, the results shows that genetic algorithm has a good solution quality more than the other Meta-heuristic algorithms for solving QAP instances.

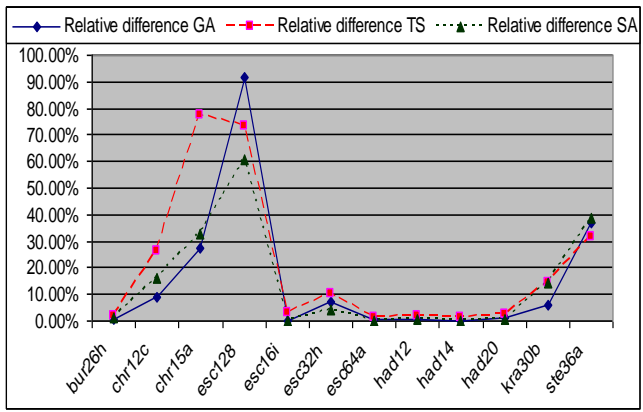

Fig. 2. Relative percentage deviation of the solution quality for different problems size

Figure.3 shows the execution time for QAP for different problems size for GA,TS and GA algorithms, the results shows that Tabu search algorithm has a faster execution time than the other Meta-heuristic algorithms for solving Real-life QAP instances.

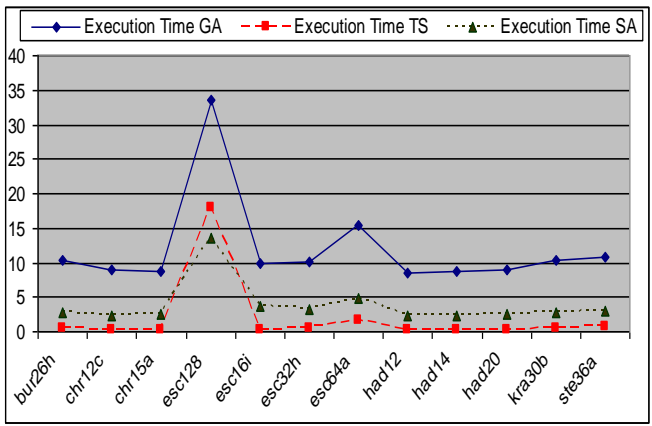

Fig. 3. Execution time for QAP for different problems size

## VI. CONCLUSION AND FUTURE WORK

In this paper, we applied Genetic algorithm (GA), tabu search (TS), and simulated annealing (SA) as Meta-heuristic algorithms for solving the Real life QAP. This research is dedicated to compare the relative percentage deviation of these solution qualities from the best known quality solution which is introduced in QAPLIB. The results show that GA, TS, and SA algorithms have effectively demonstrated the ability to solve QAP optimization problems. the computational results show that genetic algorithm has a better solution quality than the other Meta-heuristic algorithms for solving QAP problems. Tabu search algorithm has a faster execution time than the other Meta-heuristic algorithms for solving Real-life QAP problems.

In future research, comparisons between Meta-heuristic algorithms for more different types, different sizes of QAP instances and different algorithms can be conducted. Also apply Meta-heuristic algorithms to solve other combinatorial problems such as container terminals problems.

AUTHOR'S PROFILE

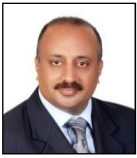

**Gamal Abd El-Nasser A. Said**: He received his M.Sc. (2012) ) in computer science from College of Computing & Information Technology, Arab Academy for Science and Technology and Maritime Transport (AASTMT), Egypt and B.Sc (1990) Faculty of Electronic Engineering, Menofia University, Egypt.

His work experience as a Researcher, Maritime Researches & Consultancies Center, Egypt. Computer Teacher, College of Technology Kingdom Of Saudi Arabia and Lecturer, Port Training Institute, (AASTMT), Egypt. Now he is Ph.D. student in computer science, Ain Shams University. His research areas include optimization, discrete-event simulation, and artificial intelligence.

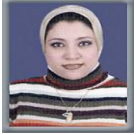

**Dr Abeer M. Mahmoud**: She received her Ph.D. (2010) in Computer science from Niigata University, Japan, her M.Sc (2004) B.Sc. (2000) in computer science from Ain Shams University, Egypt.

Her work experience is as a lecturer assistant and assistant professor, faculty, of computer and information sciences, Ain. Shams University. Her research areas include artificial intelligence medical data mining, machine learning, and robotic simulation systems.

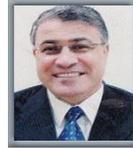

**Professor El-Sayed M. El-Horbaty**: He received his Ph.D. in Computer science from London University, U.K., his M.Sc. (1978) and B.Sc (1974) in Mathematics From Ain Shams University, Egypt. His work experience includes 39 years as an in Egypt (Ain Shams University), Qatar(Qatar University) and Emirates (Emirates University, Ajman University and ADU University). He Worked as Deputy Dean of the faculty of IT, Ajman University (2002-2008). He is working as a Vice Dean of the faculty of Computer & Information Sciences, Ain Shams University (2010-Now). Prof. El-Horbaty is current areas of research are parallel algorithms, combinatorial optimization, image processing. His work appeared in journals such as Parallel Computing, International journal of Computers and Applications (IJCA), Applied Mathematics and Computation, and International Review on Computers and software. Also he has been involved in more than 26 conferences.